\documentclass[runningheads]{llncs}
\usepackage{amsfonts}
\usepackage{amsmath}
\usepackage{booktabs}
\usepackage{graphicx}
\usepackage{tabularx}
\usepackage{makecell}
\usepackage{marvosym}
\usepackage{adjustbox}
\usepackage{multirow}
\usepackage{wrapfig}
\usepackage{lipsum}
\usepackage[table]{xcolor}
\usepackage[backref]{hyperref}
\hypersetup{
    colorlinks=true,
}

\newcommand{\mysmallscale}{\fontsize{8}{10}\selectfont}
\begin{document}
\title{Pseudo-Prompt Generating in Pre-trained Vision-Language Models for Multi-Label Medical Image Classification}
\titlerunning{Pseudo-Prompt Generating for Multi-Label Medical Images}
%
\author{Yaoqin Ye\and
Junjie Zhang\and
Hongwei Shi\textsuperscript{(\Letter)}}
\authorrunning{Y. Ye et al.}
\institute{Sichuan University, College of Computer Science, Chengdu, China\\ 
\email{cristata@stu.scu.edu.cn\\ junjiezhang1@vip.qq.com\\shihw001@126.com}}
\maketitle           

\begin{abstract}
The task of medical image recognition is notably complicated by the presence of varied and multiple pathological indications, presenting a unique challenge in multi-label classification with unseen labels. This complexity underlines the need for computer-aided diagnosis methods employing multi-label zero-shot learning. Recent advancements in pre-trained vision-language models (VLMs) have showcased notable zero-shot classification abilities on medical images. However, these methods have limitations on leveraging extensive pre-trained knowledge from broader datasets, and often depend on manual prompt construction by expert radiologists. By automating the process of prompt tuning, prompt learning techniques have emerged as an efficient way to adapt VLMs to downstream tasks. Yet, existing CoOp-based strategies fall short in performing class-specific prompts on unseen categories, limiting generalizability in fine-grained scenarios. To overcome these constraints, we introduce a novel prompt generation approach inspirited by text generation in natural language processing (NLP). Our method, named Pseudo-Prompt Generating (PsPG), capitalizes on the priori knowledge of multi-modal features. Featuring a RNN-based decoder, PsPG autoregressively generates class-tailored embedding vectors, \textit{i.e.}, pseudo-prompts. Comparative evaluations on various multi-label chest radiograph datasets affirm the superiority of our approach against leading medical vision-language and multi-label prompt learning methods. 
The source code is available at \url{https://github.com/fallingnight/PsPG}.

\end{abstract}
\section{Introduction}\label{intro}
\begin{figure}[t]
\centering
\includegraphics[width=\textwidth]{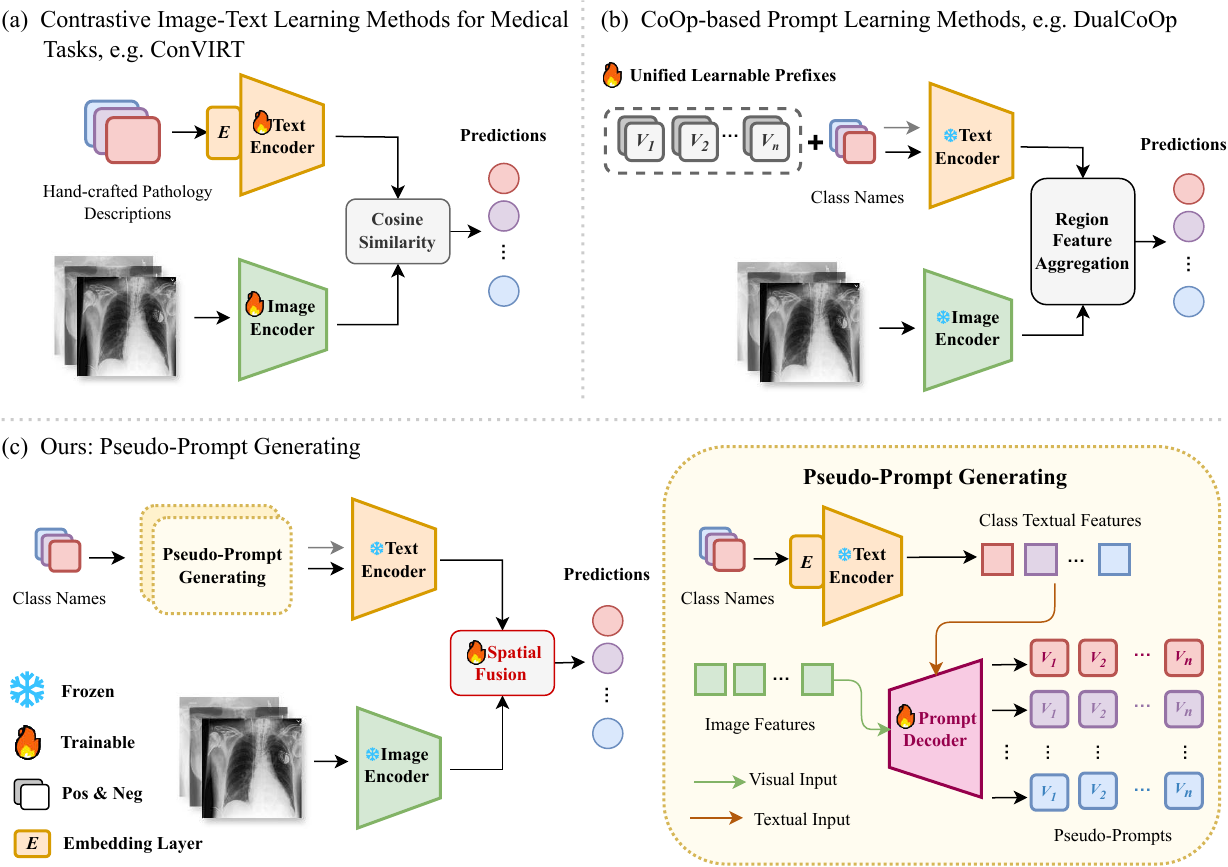}
\caption{\textbf{Comparative overview of prompt construction methods for vision-language models.} \textbf{(a)} Contrastive pretraining methods, using expert-crafted prompts\cite{gloria,convirt}; \textbf{(b)} CoOp-based methods with unified learnable prefixes\cite{dualcoop,coop}; \textbf{(c)} Our Pseudo-Prompt Generating, employing a decoder for dynamic pseudo-text generation.} \label{fig1}
\end{figure}
\footnotetext[1]{Conditions such as ``Pleural Thickening" and ``Pleural Adhesions", for instance, may be grouped into a broader ``Pleural Other" category\cite{chexpert}.}
Medical images like chest radiographs frequently contain multiple pathological features in real-world scenarios\cite{padchest,chexpert,chestxray}. Moreover, specific contexts or annotation strategies can lead to open-vocabulary pathology categorizations\footnotemark[1]. This complexity underlines the need for computer-aided diagnosis methods employing multi-label zero-shot learning to provide more reliable clinical decision-making support. Yet, the high costs associated with expert labeling of extensive chest radiograph datasets pose a barrier, resulting in a scarcity of large-scale, realistically annotated datasets. 
Despite these challenges, large medical datasets like MIMIC-CXR\cite{mimic} offer a wealth of unstructured medical report texts, enabling the exploration of self-supervised image-text contrastive learning methods\cite{gloria,CLIP,Chexzero,convirt}. These methods
develop a visual-linguistic associative latent space through extensive pre-training. During inference, they capitalize on the similarities between features of classificatory prompts and images, facilitating classification to extend to unseen categories.

However, image-text contrastive learning methods face challenges in medical applications due to the specialized knowledge required\cite{pubmedbert,kad}. Utilizing pre-training weights from large, general-purpose datasets to mitigate training costs is difficult. While we discovered that fine-tuning image-text encoder pairs with a pre-trained CLIP backbone yielded superior outcomes compared to employing a pre-trained, domain-specific BERT\cite{bioclinicalbert} as the text encoder \textbf{(CheXpert: Macro AUC 74.4\textgreater 70.9, Micro AUC 69.8\textgreater  66.7)}. Replacing the image encoder with a pre-trained Swin-T\cite{swin} echoed this trend \textbf{(Macro AUC 66.2, Micro AUC 63.5)}. We argue that, in line with insights from MaskCLIP\cite{maskclip}, utilizing encoder pairs not pre-trained via contrastive learning necessitates the re-establishment of visual-linguistic associations. Pre-training CLIP\cite{CLIP} on an extensive collection of 400M image-text pairs fosters a latent space endowed with robust generalization capabilities, whereas datasets for downstream tasks can hardly be of such a huge scale. Attempts to re-establish visual-linguistic connections will increase training costs, but hardly improve performance.

Besides, the selection of prompts critically influences classification outcomes. Traditional image-text contrastive pre-training methods rely on manually constructed prompts, a process both time-intensive and cumbersome for prompt-tuning. \cite{biovil,Chexzero} utilize generic, hand-crafted templates, whereas \cite{gloria,medclip,convirt} employ pathology descriptions provided by expert radiologist. The advent of CoOp-based prompt learning methods\cite{dualcoop,coop} marks a significant advancement by offering an automated approach for prompt tuning. Despite their simplicity and efficiency, these methods encounter limitations in generalizing to unseen categories. They tend to adopt a unified prompt template across all categories, disregarding the class-specific semantic information. Consequently, this approach often yields suboptimal results in tasks requiring fine-grained discrimination\cite{kgcoop,cocoop}.

To mitigate the challenges in deploying medical ML-ZSL models practically, our goal is to reduce computational demands without compromising on model efficacy. Addressing the previously discussed limitations of existing methods, we refine our approach by fine-tuning a pre-trained CLIP model on medical datasets and advancing the prompt learning strategy. Our method, the \textbf{Pseudo-Prompt Generating (PsPG)}, applies text generation techniques to prompt learning for the first time. Contrary to traditional approaches that target natural language texts, PsPG aims to produce high-dimensional embedding vectors, \textit{i.e.}, pseudo-prompts. We designed a decoder to dynamically generate class-tailored pseudo-prompts by tapping into the priori knowledge of multi-modal features. With the objective of generating sequences, we employ the autoregressive method to construct outputs step-by-step. Fig. \ref{fig1} contrasts our PsPG method (c) with other prompt construction methods (a) and (b) utilized in vision-language models.

In summary, our work presents these key contributions:
\begin{enumerate}
    \item We introduce the \textbf{Pseudo-Prompt Generating}, a novel application of prompt learning for multi-label zero-shot learning with medical images. PsPG uniquely generates dynamic pseudo-prompts based on multi-modal knowledge.
    \item  Leveraging visual-language model pre-training, PsPG has shown to achieve performance close to or exceeding the state-of-the-art medical vision-language model on the multi-label zero-shot classification task with significantly lower computational cost. Also, our method demonstrates a superior generalizability in multi-label prompt learning.
    \item While PsPG was designed to solve problems in multi-label zero-shot classification on medical images, its framework is versatile, offering a general solution for prompt tuning across different domains beyond healthcare.
\end{enumerate}

\section{Related Works}
\subsubsection{Multi-label Classification on Medical Images.}
In clinical practice, chest radiograph is an important disease screening tool. As we introduced in \ref{intro}, diagnosis from chest radiographs has long been a lingering issue \cite{AlbahliRAB21}, due to the complexity of those images. To address this issue, researchers proposed innovative solutions capitalized on deep learning\cite{JanizekEDL20,ZhangXPLVLSHLSX21}. However, most of them primarily focus on single-label classification. For multi-label scenarios, CheXNet\cite{chexnet} and CheXNeXt\cite{chexnext} introduced multi-label approaches using CNN. Although these supervised methods could achieve remarkable performance on seen labels, little work has focused on multi-label zero-shot learning.
\subsubsection{Vision-Language Models.}
Attempts to extract labels from radiology reports as alternatives to manual annotations through NLP \cite{chexpert,chestxray} overlook detailed descriptions and may yield noisy labels. To mitigate these, the image-text contrastive learning has propelled self-supervised knowledge extraction via zero-shot learning\cite{LarochelleEB08}. Notably, models like CLIP\cite{CLIP}, pre-trained on extensive image-text pairs, have shown exceptional skill in capturing representations. Similarly, in medical tasks, several vision-language methods leverage contrastive learning to various ends. ConVIRT\cite{convirt} exploits bi-directional targets; GLoRIA\cite{gloria} starts from global-local correlation; Chexzero\cite{Chexzero} fine-tunes CLIP; MedCLIP\cite{medclip} re-pairs images and texts by entity semantics; BioViL\cite{biovil} enhances text modeling; BioViL-T\cite{biovilt} exploits temporal correlation. Despite their advances, many depend on hand-crafted prompts. In contrast, KAD\cite{kad} reduced this dependency by employing a Disease Query Network as a classification head to foster zero-shot capabilities, though at a high computational cost.
\subsubsection{Prompt Learning.}
Prompt learning tailors pre-trained language models for specific tasks by modifying inputs instead of model parameters\cite{LiuYFJHN23}, preserving the original knowledge base while ensuring task adaptability. Context Optimization (CoOp)\cite{coop} innovates with learnable prompt vectors, and DualCoOp\cite{dualcoop} introduces dual learnable prompts for multi-label tasks. However, unified prompts falter in generalizing to unseen categories. To enhance generalization, CoCoOp\cite{cocoop} employs image-conditional prompts; KgCoOp\cite{kgcoop} balances between general and task-specific prompts; KAPT\cite{kapt} enriches prompts with Wikipedia descriptions; TCP\cite{TCP2024} utilizes textual knowledge embedding. Our approach diverges by leveraging multi-modal knowledge for more fine-grained prompt learning in multi-label medical tasks.
\section{Methodology}
\subsection{Background}\label{bg}
\subsubsection{Vison-Language Pre-training(VLP).} CLIP\cite{CLIP}, as a typical contrastive learning VLP method, its training phase is slightly different from the zero-shot inference phase. It utilizes an image encoder \(E_u\) and a text encoder \(E_w\). Given an image \(X_u\) and its corresponding text \(X_w\) , the image and text features are computed as \(u = E_u(X_u)\) and \(w = E_w(X_w)\), where \(u, w \in \mathbb{R}^{D_{vlp}}\) with \(D_{vlp}\) representing the output feature dimension. The cosine similarity between these features is computed as:
\begin{equation}
\textit{sim}(u, w) = \frac{u \cdot w}{\|u\|\|w\|}
\end{equation}
During pre-training, CLIP aims to maximize the similarity for each matched text-image pair within a mini-batch, establishing visual-linguistic associations. 

\subsubsection{Zero-shot Inference.} Let the label be \(l\), then the set of all labels is \(\mathbb{L} = \{l_1, l_2, \ldots, l_{N_c}\}\) ,where \(N_c\) is the number of categories. To perform zero-shot inference, the VLP model uses prompts \(P\) for each category, \(P\) can be denoted as \(P_i = \textit{template}(l_i)\) for \(i \in [1, N_c]\), where \(\textit{template}(cls)\) is a manually constructed template. CLIP uses \(\textit{template}(cls)\) like \textit{"a photo of a \{\(cls\)\}."} for multi-class single-label classification tasks\cite{CLIP}. Whereas some VLP models focusing on multi-label tasks use both \(\textit{template}_{pos}(cls)\) and \(\textit{template}_{neg}(cls)\) to convert multi-label tasks into separate binary tasks\cite{biovil}. The probability of each category for the input image \(X_u\) when using dual prompt can be computed as:
\begin{equation}\label{eq2}
p_i = \frac{\exp(\textit{sim}(E_u(X_u),E_w(P_{pos_i}))/\tau)}{\exp(\textit{sim}(E_u(X_u),E_w(P_{pos_i}))/\tau) + \exp(\textit{sim}(E_u(X_u),E_w(P_{neg_i}))/\tau)}
\end{equation}
where \(\tau\) refers to the temperature hyper-parameter, \(p_i\) responds to the predicted probability of \(l_i\) computed by positive and negative logits via softmax, \(i \in [1, N_c]\).

\subsubsection{Prompt Learning.} CoOp\cite{coop} introduces learnable prompts by directly learning the embedding vectors, bypassing the first embedding layer of \(E_w\). For a predefined sequence length \(n\), CoOp uses \( \textit{template}(l_i) = [V_1, V_2, \ldots, V_n, \{cls^*\}]\), with \(V_i \in \mathbb{R}^{D_{embed}}\) as learnable vectors and \(cls^*\) as the embedding vectors corresponding to \(l_i\). Like CoOp, DualCoOp\cite{dualcoop} uses dual prompts, creating separate learnable parameters for pos \& neg prompts, \textit{i.e.}, \(P_{pos}\) and \(P_{neg}\). Predictions are computed similarly as in Eq. \ref{eq2}, with first embedding layer of \(E_w\) removed.
\subsection{Pseudo-Prompt Generating}
\begin{figure}[t]
\centering
\includegraphics[width=\textwidth]{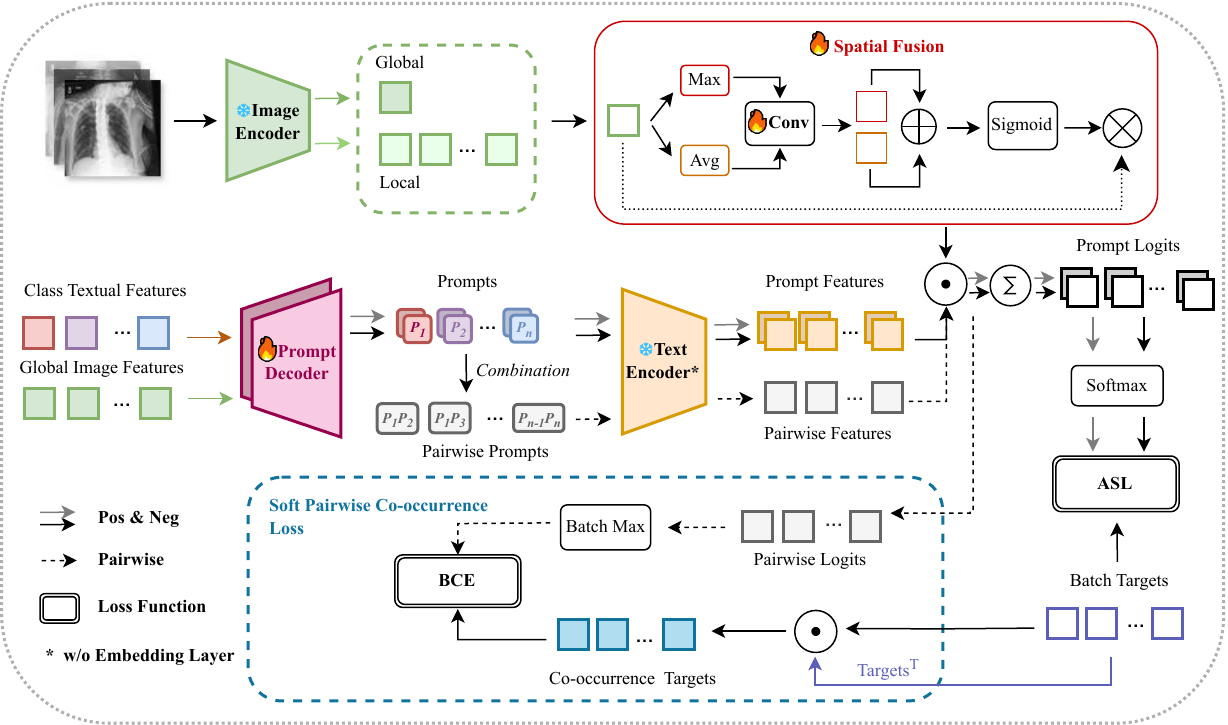}
\caption{\textbf{Overview of the prompt learning phase of PsPG.} We introduce
three novel components: \textit{Prompt Decoder}, \textit{Spatial Fusion}, and \textit{Soft Pairwise Co-occurrence Loss}.} \label{fig2}
\end{figure}
\subsubsection{Overview.} The PsPG training unfolds in two phases. Initially, it follows CLIP's pre-training to perform fine-tuning on a medical image-text dataset as in \ref{bg}. The second phase, central to our method, involves prompt learning on an image-label dataset, as depicted in Fig. \ref{fig2}. \textit{Prompt Decoder} generates class-tailored pseudo-prompts from visual global features and class textual features, illustrated in Fig. \ref{fig1} (c). Note that what we generate is not prompt templates, but complete pseudo-prompts, which do not need to be concatenated with the corresponding class names. Pseudo-prompts directly feed into the text encoder that bypasses the first embedding layer. Two prompt decoders operate in parallel to derive both positive and negative pseudo-prompts. 

Given the limitations of treating multi-label classification as binary classification tasks, we introduce a \textit{Soft Pairwise Co-occurrence Loss} \textbf{(SPCL)} alongside \textit{Asymmetric Loss} to account for label co-occurrence. Additionally, to enhance discernment of fine-grained pathologies, we apply self-attention to the image's global-local features by a spatial attention module named \textit{Spatial Fusion}.

\begin{figure}[t]
\centering
\includegraphics[width=\textwidth]{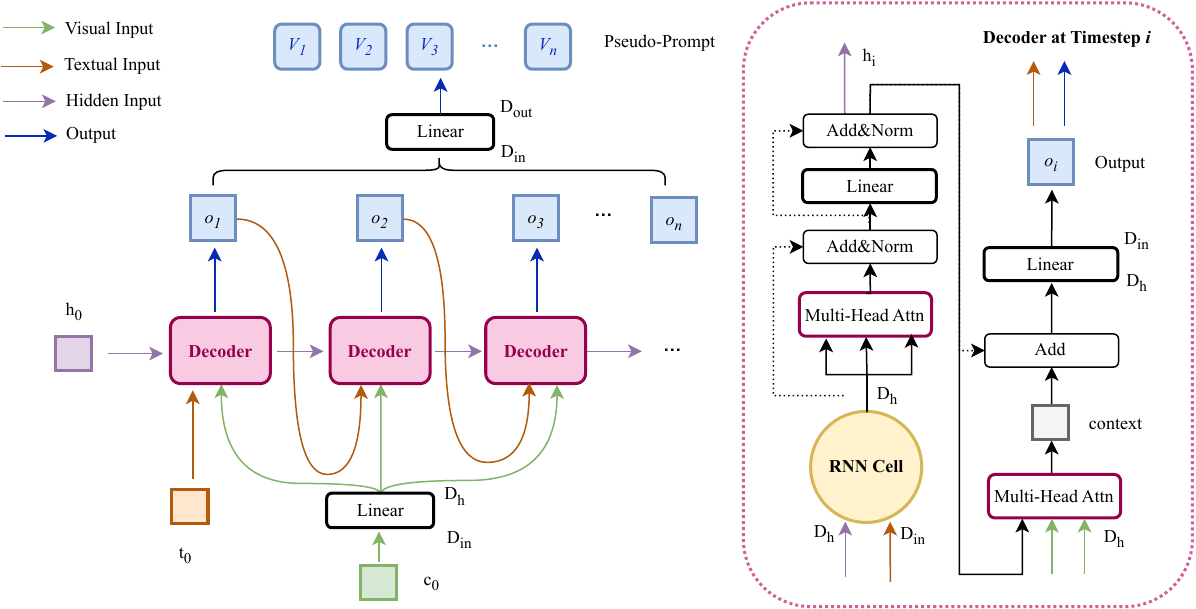}
\caption{The autoregressive process and the internal architecture of \textit{Prompt Decoder}.} \label{fig3}
\end{figure}

\subsubsection{Prompt Decoder.}\label{pd}
CoOp-based prompt learning methods\cite{dualcoop,kgcoop,coop} usually learn by directly taking the target vectors as parameters, which is a straightforward scheme. However, we observed that the learned prompts retains a sequential form, inspiring us to draw from the Seq2Seq approach\cite{rnnattn,seq2seq} prevalent in NLP tasks like dialogue systems and machine translation. In our design, we implement an autoregressive \textit{Prompt Decoder}. Unlike typical text generation that relies on natural language as the ground truth, prompt learning's supervision is indirectly derived from labels.

Fig. \ref{fig3} (left) showcases the autoregressive process of pseudo-prompt generation by \textit{Prompt Decoder}. For the identical structure of the two decoders, we just describe the process using a single example. Our method diverges from traditional Seq2Seq by utilizing image features from a mini-batch as the context rather than text encoder outputs. For a batch size \(B\) and \(N_c\) classes, we denote the global image features within the mini-batch as \(U_g = [u_1, u_2, \ldots, u_B]\). The Prompt Decoder \(G_p\) is fed with three inputs: hidden state \(h \in \mathbb{R}^{D_h}\), textual input \(t \in \mathbb{R}^{N_c \times D_{in}}\), and visual input \(c \in \mathbb{R}^{B \times D_{in}}\), where \(D_h\) is the hidden state dimension and \(D_{in} = D_{vlp}\). The initial inputs \(h_0\), \(t_0\), and \(c\) are set to all zeros, \(E_w(\mathbb{L})\), and \(E_u(U_g)\) respectively, with \(c\) duplicated \(N_c\) times to align with \(t_0\) and transformed to \(c' \in \mathbb{R}^{N_c \times B \times D_h}\) through a linear layer. At each time-step \(i\), the decoder outputs \(o_i \in \mathbb{R}^{D_{in}}\) which serves as the input \(t\) for the next time-step \(i+1\), computed as:
\begin{equation}
o_i = G_p(h_{i-1}, o_{i-1}, c')
\end{equation}
This yields the outputs \(O = [o_1, o_2, \ldots, o_n]\). In CLIP, since \(D_{embed} \neq D_{vlp}\), \(D_{out} \neq D_{in}\), the output \(O\) should be projected into \(\mathbb{R}^{D_{out}}\) to form the final prompts \(P = [V_1, V_2, \ldots, V_n]\). Our prompt learning's sole aim is to enhance VLP classification, eliminating the need to revert to natural language forms during inference, thus the output is solely high-dimensional embedding vectors. Consequently, the pseudo-prompt length is fixed to \(n\).

Fig. \ref{fig3} (right) details the decoder's internal architecture. We employ an RNN Cell \(g\) implemented by GRU for the autoregressive mechanism. At time-step \(i\), the RNN Cell outputs \(h'_i = g(t_{i-1}, h_{i-1})\). We utilize a self-attention pipeline like Transformer\cite{transformer} to refine \(h'_i\), resulting in \(h_i = \text{Self-Attn}(h'_i, h'_i, h'_i)\) where \(q, k, v = h'_i\). For the visual inputs \(c'\), we use cross-attention to compute the context:
\begin{equation}
\textit{context} = \text{Attn}(h_i, c', c')
\end{equation}
where \(q = h_i\), \(k = c'\), \(v = c'\), and \(\text{Attn}\) is a multi-head attention layer. For simplicity, we omit the permute and resize steps.

Lastly, we apply a residual concatenation to derive \(o'_i = h_i + \textit{context}\). Since \(o'_i\) will be linearly projected to \(o_i\), we do not use additional batch normalization.

\subsubsection{Global \& Local Spatial Fusion.}
The presence of fine-grained pathological indications in chest radiographs necessitates an approach to focus on localized nuance of images\cite{biovil,gloria}. We extract local features \(U_l \in \mathbb{R}^{C_{local} \times D_{vlp}}\) from the image encoder's last pooling layer and global features \(U_g \in \mathbb{R}^{1 \times D_{vlp}}\) from the entire encoder's output, where \(C_{local}\) is the number of channels for local outputs.

We introduce a learnable lightweight spatial attention module \(SF\), named \textit{Spatial Fusion}, with only 3 parameters in a 1D convolution as shown in Fig. \ref{fig2}. This module is adapted from ECA\cite{ECA}, but used for spatial self-attention and consider both maximum pooling and average pooling. The fused global-local feature is computed by \(U_{gl}  = SF(\text{concat}(U_g, U_l))\), where \(U_{gl} \in \mathbb{R}^{(C_{local}+1) \times D_{vlp}}\).

This approach contrasts with the parameter-free attention in DualCoOp\cite{dualcoop}. Ours focuses on the global-local features rather than logits after image-text projection.
\subsubsection{Soft Pairwise Co-occurrence Loss.}
The PsPG loss function is composed of \textit{Asymmetric Loss} (ASL)\cite{ASL} as \(L_\text{ASL}\) for addressing category imbalance in multi-label medical datasets and \textit{Soft Pairwise Co-occurrence Loss} \textbf{(SPCL)} to capture the label correlation. 

\textbf{SPCL} is inspired by the \textit{Pairwise Label Co-occurrence Prediction} \textbf{(PLCP)} task\cite{ZhangZYLC21} in multi-label text classification.  
We consider all label co-occurrences within the mini-batch and convert label pair co-occurrence prediction to a multi-label classification task with \(C_2^{N_c}\) classes.
\(A = [a_1, a_1, ...,a_B]\) is denoted as batch-level labels, where \(a_i\) is the label of image \(X_{u_i}\), \(A \in \{0,1\}^{N_c \times B}\). We can simply compute the batch-level labels co-occurrence matrix \(\mathrm{\Omega}\):
\begin{equation}
\mathrm{\Omega} = A^TA
\end{equation}
where \(\mathrm{\Omega} \in \mathbb{Z}^{N_c \times N_c}\). \(\mathrm{\Omega}\) is a symmetric matrix, and the element \(\omega_{ij}\) of \(\mathrm{\Omega } ( i \in [1,Nc], j \in [1,Nc], i<j)\) denotes the times of co-occurrences of label \(l_i\) and \(l_j\) within the mini-batch. For simplicity, we construct co-occurrence targets \(Q\) by flattening the strictly upper triangular matrix of  \(\mathrm{\Omega}\) and setting all \(\omega_{ij} > 1\) to 1, \(Q \in \{0,1\}^{C_2^{N_c}}\).

Fig. \ref{fig2} illustrates that using the pairwise concatenation of pseudo-prompts \(P_iP_j\) (where \(i, j \in [1, N_c]\) and \(i < j\)) to yield \(W_{pair}\), with \(W_{pair} \in \mathbb{R}^{C_2^{N_c} \times D_{vlp}}\). For the global-local feature \(U_{gl}\) of a batch, the probability \(p_{mn}\) is given by:
\begin{equation}
p_{mn} = \frac{1}{1 + \exp\left(-\text{sim}(u_{gl}, w_{ij})/\tau\right)}
\end{equation}
where \(\tau\) shared with non-pairwise prompts, \(m \in [1, C_2^{N_c}]\), \(n \in [1, B]\), and \(w_{ij} \in W_{\text{pair}}\).

The maximum pooled probability \(\text{pool}_{p_i}\) across the batch and the pairwise loss \(L_{\text{SPCL}}\) are computed as:
\begin{equation}
\text{pool}_{p_i} = \max(p_{i1}, p_{i2}, \ldots, p_{iB})
\end{equation} 
\begin{equation}
L_{\text{SPCL}} = \sum_{i=1}^{C_2^{N_c}} \left[q_i \log(\text{pool}_{p_i}) + (1 - q_i) \log(1 - \text{pool}_{p_i})\right]
\end{equation}

We use the batch-level co-occurrence target \(Q\) to serve as the soft target for all label pair co-occurrence probabilities within the mini-batch. Maximum pooling is applied to these probabilities to computing the \textit{Binary Cross-Entropy Loss} with the soft target \(Q\). 
The overall loss \(L\) for PsPG is:
\begin{equation}
L = L_{\text{ASL}} + L_{\text{SPCL}}
\end{equation}

\section{Experiments}
\subsection{Setup}
\subsubsection{Datasets.}We use MIMIC-CXR\cite{mimic} as the train set in the fine-tuning phase and CheXpert-train\cite{chexpert} in the prompt learning phase. We evaluated the multi-label performance against baselines using four datasets: CheXpert-test, PadChest\cite{padchest}, VinDr-CXR\cite{vindr}, and Private-CXR. For all datasets containing multiple views, we selected all frontal images based on metadata.

\textit{Private-CXR} is a chest radiograph dataset collected from 
The First People's Hospital of Hangzhou Lin'an District, containing 7,357 images and 18 pathology categories all labelled by experienced radiologists. The availability of this dataset is restricted. Contingent on participant consent, we used the entire dataset for zero-shot evaluation. 

\subsubsection{Baselines.}\textit{Medical Vision-Language Pre-training Methods:} These pioneering contrastive learning methods, include ConVIRT\cite{convirt}, GLoRIA\cite{gloria}, BioViL-T\cite{biovilt}, MedCLIP\cite{medclip}, Chexzero\cite{Chexzero} and KAD\cite{kad}, were known for their efficacy in zero-shot medical image classification\footnotemark[2]. Pre-trained weights were utilized where available. For which absent such weights\cite{convirt}, we retrained it on MIMIC-CXR. Additionally, we compared with the fine-tuned CLIP prior to prompt tuning.

\noindent\textit{Prompt Learning Methods:} We compared PsPG with multi-label prompt learning method: DualCoOp\cite{dualcoop}. We aligned it with PsPG in using the same CLIP backbone and underwent supervised training on CheXpert. 
\footnotetext[2]{We extended the originally single-label focused to multi-label, as detailed in \ref{bg}.}
\subsubsection{Train.}Aligning with most baselines, we used ResNet50 backbone weights provided by CLIP\cite{CLIP}. In backbone fine-tuning, we use AdamW optimizer with weight decay 0.05, and set learning rate to \(1 \times 10^{-4}\) with cosine annealing. We trained 15 epochs, with batch size 64, 5 warm-up epochs and warm-up learning rate of \(1 \times 10^{-6}\). With a frozen backbone, we tuned the prompt decoder (sequence length 16, hidden dimension 512) and set ASL \(\gamma^+ = 1\), \(\gamma^- = 2\), and \(c = 0.05\). We use SGD optimizer, and set learning rate to \(1 \times 10^{-4}\) with cosine annealing. We trained 50 epochs, with batch size 64, 5 warm-up epochs and warm-up learning rate of \(1 \times 10^{-6}\). We select the best epoch by evaluating the Macro AUC performance on the CheXpert-val. All training phases are done with one NVIDIA RTX 3080Ti.
\subsubsection{Evaluation.} For PsPG, KAD, and DualCoOp, we used class names as input. For other baselines, we applied fixed templates from BioViL\footnotemark[3]. Similar to typical medical tasks and multi-label tasks, we use Macro AUC, Micro AUC, mAP as evaluation metrics.
Predictions confidence intervals were computed using a non-parametric bootstrap with 1000 re-samples and \(\alpha\) of 0.05. Due to page limitations, we placed the detailed confidence intervals in appendix. 
\footnotetext[3]{Positive: ``Findings suggesting \{cls\}." Negative: ``No evidence of \{cls\}."}
\subsection{Multi-label Zero-shot Classification} \label{mlzc}
To make evaluation categories consistent, we evaluated Generalized Zero-Shot Learning (GZSL) on four datasets. Unlike the CheXpert original study's focus on five competition categories\cite{chexpert}, Our evaluation spans all 14 pathology categories. In Table \ref{table1}, \(\text{CLIP}_{\text{fine-tuned}}\) denotes the CLIP we fine-tune on the medical dataset. \(\text{DualCoOp}_{\text{init}}\) denotes template initialization using \footnotemark[3], and \(\text{DualCoOp}_{\text{16}}\) denotes using the same length of prompt sequences as ours. \(\textbf{PsPG}_{\text{prefix}}\) denotes using our method to learn prefixes than whole prompts.  Additionally, we compare the performance of fine-tune the pre-trained CheXpert weights on ChestXray-14\cite{chestxray} for \(\text{DualCoOp}_{\text{cross}}\) and \(\textbf{PsPG}_{\text{cross}}\). 
\begin{table}[ht]
\centering
\caption{\textbf{Multi-label Zero-shot Classification on CheXpert, VinDr-CXR, Private-CXR, PadChest.} \#P stands for trainable parameters of modules. All metrics are displayed in percentage. For PadChest, we only evaluate on categories which positive samples \(n_{pos} > 50\) (108 categories). We underlined the best performance across all methods (except cross methods).All models except BioViL-T* (only provided \(512 \times 512\)) evaluated used \(224 \times 224\) as visual input.}

\label{table1}
\resizebox{\textwidth}{!}{
\begin{tabular}{@{}lccccccccccccccccc@{}}
\toprule
    \textbf{Methods} & \textbf{\#P} & \multicolumn{4}{c}{\textbf{CheXpert}\cite{chexpert}} & \multicolumn{4}{c}{\textbf{VinDr-CXR}\cite{vindr}} & \multicolumn{4}{c}{\textbf{Private-CXR}} & \multicolumn{4}{c}{\textbf{PadChest}\cite{padchest}}\\
\cmidrule(lr){3-6} \cmidrule(lr){7-10}\cmidrule(lr){11-14} \cmidrule(lr){15-18}
 &  & \mysmallscale{\makecell{Macro\\AUC}} & \mysmallscale{\makecell{Micro\\AUC}} & \mysmallscale{mAP} &  &  \mysmallscale{\makecell{Macro\\AUC}} & \mysmallscale{\makecell{Micro\\AUC}} & \mysmallscale{mAP} &  &  \mysmallscale{\makecell{Macro\\AUC}} & \mysmallscale{\makecell{Micro\\AUC}} & \mysmallscale{mAP}&  & \mysmallscale{\makecell{Macro\\AUC}} & \mysmallscale{\makecell{Micro\\AUC}} & \mysmallscale{mAP} &  \\
\midrule
ConVIRT\cite{convirt}      & 138.05M     & 70.7 & 66.2 & 30.3 &  & 72.7 & 71.7 & 13.0 &  & 70.6 & 64.3 & 11.6 & & 63.7 & 58.5 & 3.6 &\\
GLoRIA\cite{gloria} & 134.18M & 75.9 & 73.2 & 42.3 &  & 64.2 & 51.9 & 13.8 &  & 76.9 & 56.7 & 16.2 & & 67.3 & 59.3 & 4.5 &\\
MedCLIP\cite{medclip} & 136.62M & 76.6 & 83.4 & 47.7 &  & 58.9 & 71.6 & 22.4 &  & 57.2 & 71.1 & 18.0 & & 54.7 & 54.5 & 5.3 &\\
BioViL-T*\cite{biovilt}  & 136.98M & 70.6 & 67.6 & 36.8 &  & 81.2 & 72.2 & 19.3 &  & 76.0 & 61.6 & 14.8 & & 69.2 & 66.9 & 5.3 &\\
Chexzero\cite{Chexzero}  & 151.28M & 72.7 & 60.6 & 41.6 &  & 75.9 & 62.6 & 19.6 &  & 72.1 & 71.1 & 18.6 & & 65.8 & 65.7 & 5.9 &\\
\(\text{CLIP}_{\text{fine-tuned}}\)\cite{CLIP} & 102.01M & 74.4 & 69.8 & 42.0 &  & 75.0 & 71.8 & 18.5 &  & 76.3 & 72.4 & 16.5 & & 69.9 & 67.2 & 6.4 &\\ 
KAD\cite{kad}  &  \textbf{163.28M} & 83.8 & 85.2 & \underline{50.6} &  & \underline{82.6} & \underline{82.4} & \underline{24.7} &  & 77.8 & 70.1 & \underline{20.7} & & \underline{75.7} & \underline{79.1} & \underline{10.1} &\\
\midrule
\rowcolor{pink!20}
\(\text{DualCoOp}_{\text{init}}\)\cite{dualcoop}& 0.02M & 83.6 & 84.6 & 46.3 &  & 72.9 & 74.7 & 19.6 &  & 67.5 & 51.1 & \textbf{19.6} & & \textbf{65.7} & \textbf{67.1} & \textbf{7.3}&\\
\rowcolor{pink!20}
\(\text{DualCoOp}_{\text{16}}\)\cite{dualcoop} & 0.02M & 85.0 & \textbf{\underline{85.4}} & 47.5 &  & 72.2 & 69.8 & 18.6 &  & 69.9 & 58.0 & 18.5 & & 64.0 & 63.5 & 6.1 &\\
\rowcolor{pink!20}
\(\textbf{PsPG}_{\text{prefix}}\) & 12.60M & 84.8 & 84.8 & 46.4 &  & 67.8 & 75.2 & 18.2 &  & 67.7 & 66.9 & 18.4 & & 60.8 & 63.0 & 4.3 &\\
\rowcolor{pink!20}
\textbf{PsPG} & 12.60M & \textbf{\underline{85.2}} & 85.1 & \textbf{49.3} &  & \textbf{75.1} & \textbf{78.4} & \textbf{19.9} &  &  \textbf{\underline{77.9}} & \textbf{\underline{80.5}} & 18.8 & & 62.2 & 61.6 & 4.0 &\\
\midrule
\rowcolor{yellow!10}
\(\text{DualCoOp}_{\text{cross}}\)\cite{dualcoop} & 0.02M & 78.4 & \textbf{76.3} & 41.8 &  & 75.5 & 78.9 & 19.3 &  & 68.3 & 67.4 & \textbf{17.9} & & 65.3 & \textbf{67.6} & \textbf{7.2} &\\
\rowcolor{yellow!10}
\(\textbf{PsPG}_{\text{cross}}\) & 12.60M & \textbf{82.8} & 75.6 & \textbf{45.2} &  & \textbf{82.4} & \textbf{81.5} & \textbf{22.3} &  & \textbf{77.1} & \textbf{81.0} & 17.1 & & \textbf{66.5} & 61.7 & 4.3 &\\
\bottomrule
\end{tabular}
}
\end{table}
\subsubsection{Result.}
Our PsPG excelled in zero-shot tasks across datasets. On CheXpert, it is noteworthy that prompt learning methods excel over most contrastive pre-training ones. We led in all Macro AUC and mAP expect KAD. While KAD, the current state-of-the-art pre-training method, demonstrates strong generalizability, it incurs high deployment costs, consuming \textgreater 22.5GiB GPU memories for evaluating PadChest with 108 categories. Our method, with less than a tenth of trainable parameters
\footnotemark[4]
, achieves comparable or superior performance. 
\footnotetext[4]{\textless 3.1GiB of GPU memories were consumed when evaluating PadChest.}

On VinDr-CXR, we surpassed DualCoOp and most pre-training methods. And in some Private-CXR metrics, we outperformed KAD (Macro AUC +0.1, Micro AUC +10.4, mAP -1.9). The excellent performance of micro AUC highlight our method's robustness when evaluating overall data, despite weaker generalization on the on the Spanish dataset PadChest compared to \(\text{CLIP}_{\text{fine-tuned}}\) and DualCoOp. That reminds us that although we try to exploit multi-modal prior knowledge, it is difficult to generalize to datasets with 108 categories (102 unseen) when the training label semantic space is narrow. We believe that a better approach would be to train on datasets with a wide range of categories. 

Fine-tuning with Chestxray-14 underscores our method's adaptability. Our gains on VinDr-CXR (Macro AUC +7.3, -0.2 KAD, +1.2 BioViL-T; Micro AUC +3.1, -0.9 KAD; mAP +2.4, -2.4 KAD, -0.1 MedCLIP) and a better Micro AUC (+0.5) on Private-CXR demonstrate potential for broad dataset applications.

Attempts to optimize by setting the decode target to prefix \(\textbf{PsPG}_{\text{prefix}}\) did not enhance zero-shot task performance. We speculate that this operation reuses label semantic information, which reinforces our dependence on the label semantic space seen during training, limiting the generalizability to unseen categories.
\subsection{Ablation Study}
\noindent\textbf{Loss Ablation.}
In Table \ref{table3}, SPCL is our \textit{Soft Pairwise Co-occurrence Loss}. While PCL is SPCL use sample-level co-occurrence. SPCL's computation of \( C_2^{N_c}\) prompts can be intensive for large category sets, which motivates trials with classic \textit{Ranking Loss}\cite{ranking} for multi-label classification. Although fitting capabilities was comparable across different losses, we observed notable disparities in zero-shot generalization. SPCL improved upon PCL by leveraging batch-level co-occurrence supervision to alleviate the rigidity of sample-level co-occurrence constraints. \textit{Ranking Loss} also proved effective, particularly enhancing Macro AUC, presenting a viable alternative in our work.

\begin{wraptable}{r}{0.5\textwidth} 

\centering
\resizebox{0.5\textwidth}{!}{
\begin{tabular}{@{}lcccccccccc@{}}
\toprule
    {\mysmallscale{\textbf{Loss}}}  & \multicolumn{3}{c}{\mysmallscale{\textbf{CheXpert}}} & \multicolumn{3}{c}{\mysmallscale{\textbf{VinDr}}} & \multicolumn{3}{c}{\mysmallscale{\textbf{Private}}} &\\
\midrule
 \mysmallscale{BCE} & 84.2 & 84.5 &  & 74.2 & 75.3&  & 72.6 & 74.6&\\
 \mysmallscale{ASL} & 83.6 & 84.9 &  & 73.1 & 70.5 &  & 68.7 & 62.0 &\\
  \mysmallscale{ASL+Ranking} & 85.6 & 85.4 &  & \textbf{78.4} & 72.1 &  & 76.2 & 74.5 &\\
\mysmallscale{ASL+PCL} & 85.4 & 86.1&  & 68.7 & 76.1&  & 67.8 & 71.4 &\\
\mysmallscale{ASL+SPCL}& 85.2 & 85.1 &  & 75.1 & \textbf{78.4}&  & \textbf{77.9} & \textbf{80.5}&\\

\toprule
    {\mysmallscale{\textbf{Features}}}  & \multicolumn{3}{c}{\mysmallscale{\textbf{CheXpert}}} & \multicolumn{3}{c}{\mysmallscale{\textbf{VinDr}}} & \multicolumn{3}{c}{\mysmallscale{\textbf{Private}}} &\\
\midrule
 \mysmallscale{G} & 84.2 & 86.0 &  & 73.7 & 77.6&  & 71.2 & 73.0&\\
 \mysmallscale{G \& L} & 84.1 & 84.5 &  & 70.9 & 76.8 &  & 71.3 & 64.2 &\\
\mysmallscale{G \& L w/ SF}& 85.2 & 85.1 &  & \textbf{75.1} & \textbf{78.4}&  & \textbf{77.9} & \textbf{80.5}&\\
\bottomrule
\end{tabular}}
\caption{This table includes Loss Ablation and Spatial Fusion Ablation. For Each Dataset, 1st column is Macro AUC and 2nd column is Micro AUC.}
\label{table3}

\end{wraptable}

\noindent\textbf{Spatial Fusion Ablation.}\label{sfa}
In Table. \ref{table3}, G denotes using only global features as original CLIP. G \& L denotes using both global and local features but simply sum the logits. We found that merely combining global and local features underperformed using solely global features in zero-shot tasks, underscoring the importance of the Spatial Fusion module.

\noindent\textbf{Decoder Ablation.}

\noindent\textit{Architecture Ablation:} We evaluated an RNN (GRU) decoder against alternative architectures \ref{table4}: a Transformer decoder (width 512, \#P 19.44M) and a non-autoregressive (NAR) decoder (as the knowledge extractor in TCP\cite{TCP2024}, width 512, \#P 9.45M). The NAR decoder showed marginally better fitting but significantly poorer zero-shot performance. The Transformer's effectiveness seemed to be constrained by the limited semantic range of our labels.

\noindent\textit{Prompt Ablation:} We evaluated three prompt generation strategies \ref{table4}: single positive prompt generation
, dual prompt generation with one decoder, and dual prompt generation with two independent decoders. The latter proved most effective. The inferior fitting of single positive prompts may stem from the lack of negative information.
\begin{table}[bh]
\centering

\caption{This table includes Decoder Architecture Ablation and Prompt Ablation. We present Macro AUC and Micro AUC.}
\label{table4}
\resizebox{\textwidth}{!}{
\begin{tabular}{@{}lccccccccc|lccccccccccc@{}}
\toprule
    {\mysmallscale{\textbf{Arch.}}}  & \multicolumn{3}{c}{\mysmallscale{\textbf{CheXpert}}} & \multicolumn{3}{c}{\mysmallscale{\textbf{VinDr}}} & \multicolumn{3}{c|}{\mysmallscale{\textbf{Private}}} & {\mysmallscale{\textbf{ Prompt}}}  & \multicolumn{3}{c}{\mysmallscale{\textbf{CheXpert}}} & \multicolumn{3}{c}{\mysmallscale{\textbf{VinDr}}} & \multicolumn{3}{c}{\mysmallscale{\textbf{Private}}} &\\
\midrule
 \mysmallscale{NAR\cite{TCP2024}} & 84.9 & 85.9 &  & 58.2 & 64.1&  & 60.5 & 61.1& &\mysmallscale{ Pos Only} & \underline{79.7} & 83.6 &  & 72.7 & 71.9&  & 67.5 & 75.1&\\
 \mysmallscale{Transformer} & 83.4 & 84.5 &  & 72.0 & 74.9 &  & 71.3& 69.5 & &\mysmallscale{ Single Decoder} & 84.3 & 84.5 &  & 69.0 & 74.0&  & 71.8 & 64.4&\\
\mysmallscale{RNN}& 85.2 & 85.1 &  & \textbf{75.1} & \textbf{78.4} &  & \textbf{77.9} & \textbf{80.5}&  & \mysmallscale{ Dual Decoder} & 85.2 & 85.1 &  & \textbf{75.1} & \textbf{78.4} &  & \textbf{77.9} & \textbf{80.5}&\\
\bottomrule
\end{tabular}}

\end{table}

\section{Conclusion}
In this work, we introduce \textbf{Pseudo-Prompt Generating}, showing a promising direction for multi-label zero-shot learning with medical images. With a compelling blend of efficiency and efficacy, PsPG autoregressively generates class-tailored pseudo-prompts, leveraging \textit{Spatial Fusion} and \textit{Soft Pairwise Co-occurrence Loss} for enhanced multi-label performance. Extensive experiments proves that it matches or even surpasses state-of-the-art models, with robust generalization abilities and reduced computational demands. The effectiveness of PsPG suggests the potential for its broader application across domains.

%
%
%
\bibliographystyle{splncs04}
\bibliography{export}

\begin{thebibliography}{10}
\providecommand{\url}[1]{\texttt{#1}}
\providecommand{\urlprefix}{URL }
\providecommand{\doi}[1]{https://doi.org/#1}

\bibitem{AlbahliRAB21}
Albahli, S., Rauf, H.T., Algosaibi, A.A., Balas, V.E.: Ai-driven deep {CNN} approach for multi-label pathology classification using chest x-rays. PeerJ Comput. Sci.  \textbf{7}, ~e495 (2021)

\bibitem{bioclinicalbert}
Alsentzer, E., Murphy, J., Boag, W., Weng, W.H., Jin, D., Naumann, T., McDermott, M.: Publicly available clinical bert embeddings. In: Proceedings of the 2nd Clinical Natural Language Processing Workshop. pp. 72--78. Association for Computational Linguistics (2019)

\bibitem{rnnattn}
Bahdanau, D., Cho, K., Bengio, Y.: Neural machine translation by jointly learning to align and translate. In: {ICLR} (2015)

\bibitem{biovilt}
Bannur, S., Hyland, S.L., Liu, Q., P{\'{e}}rez{-}Garc{\'{\i}}a, F., Ilse, M., Castro, D.C., Boecking, B., Sharma, H., Bouzid, K., Thieme, A., Schwaighofer, A., Wetscherek, M., Lungren, M.P., Nori, A.V., Alvarez{-}Valle, J., Oktay, O.: Learning to exploit temporal structure for biomedical vision-language processing. In: {CVPR}. pp. 15016--15027. {IEEE} (2023)

\bibitem{biovil}
Boecking, B., Usuyama, N., Bannur, S., Castro, D.C., Schwaighofer, A., Hyland, S.L., Wetscherek, M., Naumann, T., Nori, A.V., Alvarez{-}Valle, J., Poon, H., Oktay, O.: Making the most of text semantics to improve biomedical vision-language processing. In: {ECCV} {(36)}. Lecture Notes in Computer Science, vol. 13696, pp. 1--21. Springer (2022)

\bibitem{padchest}
Bustos, A., Pertusa, A., Salinas, J.M., de~la Iglesia{-}Vay{\'{a}}, M.: Padchest: {A} large chest x-ray image dataset with multi-label annotated reports. Medical Image Anal.  \textbf{66},  101797 (2020)

\bibitem{pubmedbert}
Gu, Y., Tinn, R., Cheng, H., Lucas, M., Usuyama, N., Liu, X., Naumann, T., Gao, J., Poon, H.: Domain-specific language model pretraining for biomedical natural language processing. {ACM} Trans. Comput. Heal.  \textbf{3}(1),  2:1--2:23 (2022)

\bibitem{gloria}
Huang, S., Shen, L., Lungren, M.P., Yeung, S.: Gloria: {A} multimodal global-local representation learning framework for label-efficient medical image recognition. In: {ICCV}. pp. 3922--3931. {IEEE} (2021)

\bibitem{chexpert}
Irvin, J., Rajpurkar, P., Ko, M., Yu, Y., Ciurea{-}Ilcus, S., Chute, C., Marklund, H., Haghgoo, B., Ball, R.L., Shpanskaya, K.S., Seekins, J., Mong, D.A., Halabi, S.S., Sandberg, J.K., Jones, R., Larson, D.B., Langlotz, C.P., Patel, B.N., Lungren, M.P., Ng, A.Y.: Chexpert: {A} large chest radiograph dataset with uncertainty labels and expert comparison. In: {AAAI}. pp. 590--597. {AAAI} Press (2019)

\bibitem{JanizekEDL20}
Janizek, J.D., Erion, G.G., DeGrave, A.J., Lee, S.: An adversarial approach for the robust classification of pneumonia from chest radiographs. In: {CHIL}. pp. 69--79. {ACM} (2020)

\bibitem{mimic}
Johnson, A.E.W., Pollard, T.J., Berkowitz, S.J., Greenbaum, N.R., Lungren, M.P., ying Deng, C., Mark, R.G., Horng, S.: Mimic-cxr, a de-identified publicly available database of chest radiographs with free-text reports  \textbf{6} (2019)

\bibitem{kapt}
Kan, B., Wang, T., Lu, W., Zhen, X., Guan, W., Zheng, F.: Knowledge-aware prompt tuning for generalizable vision-language models. In: {ICCV}. pp. 15624--15634. {IEEE} (2023)

\bibitem{LarochelleEB08}
Larochelle, H., Erhan, D., Bengio, Y.: Zero-data learning of new tasks. In: {AAAI}. pp. 646--651. {AAAI} Press (2008)

\bibitem{LiuYFJHN23}
Liu, P., Yuan, W., Fu, J., Jiang, Z., Hayashi, H., Neubig, G.: Pre-train, prompt, and predict: {A} systematic survey of prompting methods in natural language processing. {ACM} Comput. Surv.  \textbf{55}(9),  195:1--195:35 (2023)

\bibitem{swin}
Liu, Z., Lin, Y., Cao, Y., Hu, H., Wei, Y., Zhang, Z., Lin, S., Guo, B.: Swin transformer: Hierarchical vision transformer using shifted windows. In: {ICCV}. pp. 9992--10002. {IEEE} (2021)

\bibitem{vindr}
Nguyen, H.Q., Lam, K., Le, L.T., Pham, H.H., Tran, D.Q., Nguyen, D.B., Le, D.D., Pham, C.M., Tong, H.T.T., Dinh, D.H., Do, C.D., Doan, L.T., Nguyen, C.N., Nguyen, B.T., Nguyen, Q.V., Hoang, A.D., Phan, H.N., Nguyen, A.T., Ho, P.H., Ngo, D.T., Nguyen, N.T., Nguyen, N.T., Dao, M., Vu, V.: Vindr-cxr: An open dataset of chest x-rays with radiologist's annotations  \textbf{9} (2022)

\bibitem{CLIP}
Radford, A., Kim, J.W., Hallacy, C., Ramesh, A., Goh, G., Agarwal, S., Sastry, G., Askell, A., Mishkin, P., Clark, J., Krueger, G., Sutskever, I.: Learning transferable visual models from natural language supervision. In: {ICML}. Proceedings of Machine Learning Research, vol.~139, pp. 8748--8763. {PMLR} (2021)

\bibitem{chexnext}
Rajpurkar, P., Irvin, J., Ball, R.L., Zhu, K., Yang, B., Mehta, H., Duan, T., Ding, D., Bagul, A., Langlotz, C.P., Patel, B.N., Yeom, K.W., Shpanskaya, K., Blankenberg, F.G., Seekins, J., Amrhein, T.J., Mong, D.A., Halabi, S.S., Zucker, E.J., Ng, A.Y., Lungren, M.P.: Deep learning for chest radiograph diagnosis: A retrospective comparison of the chexnext algorithm to practicing radiologists  \textbf{15} (2018)

\bibitem{chexnet}
Rajpurkar, P., Irvin, J., Zhu, K., Yang, B., Mehta, H., Duan, T., Ding, D.Y., Bagul, A., Langlotz, C.P., Shpanskaya, K.S., Lungren, M.P., Ng, A.Y.: Chexnet: Radiologist-level pneumonia detection on chest x-rays with deep learning. CoRR  \textbf{abs/1711.05225} (2017)

\bibitem{ASL}
Ridnik, T., Baruch, E.B., Zamir, N., Noy, A., Friedman, I., Protter, M., Zelnik{-}Manor, L.: Asymmetric loss for multi-label classification. In: {ICCV}. pp. 82--91. {IEEE} (2021)

\bibitem{dualcoop}
Sun, X., Hu, P., Saenko, K.: Dualcoop: Fast adaptation to multi-label recognition with limited annotations. In: NeurIPS (2022)

\bibitem{seq2seq}
Sutskever, I., Vinyals, O., Le, Q.V.: Sequence to sequence learning with neural networks. In: {NIPS}. pp. 3104--3112 (2014)

\bibitem{Chexzero}
Tiu, E., Talius, E., Patel, P., Langlotz, C.P., Ng, A.Y., Rajpurkar, P.: Expert-level detection of pathologies from unannotated chest x-ray images via self-supervised learning  \textbf{6},  1399--1406 (2022)

\bibitem{transformer}
Vaswani, A., Shazeer, N., Parmar, N., Uszkoreit, J., Jones, L., Gomez, A.N., Kaiser, L., Polosukhin, I.: Attention is all you need. In: {NIPS}. pp. 5998--6008 (2017)

\bibitem{ranking}
Wang, J., Song, Y., Leung, T., Rosenberg, C., Wang, J., Philbin, J., Chen, B., Wu, Y.: Learning fine-grained image similarity with deep ranking. In: {CVPR}. pp. 1386--1393. {IEEE} Computer Society (2014)

\bibitem{ECA}
Wang, Q., Wu, B., Zhu, P., Li, P., Zuo, W., Hu, Q.: Eca-net: Efficient channel attention for deep convolutional neural networks. In: {CVPR}. pp. 11531--11539. Computer Vision Foundation / {IEEE} (2020)

\bibitem{chestxray}
Wang, X., Peng, Y., Lu, L., Lu, Z., Bagheri, M., Summers, R.M.: Chestx-ray8: Hospital-scale chest x-ray database and benchmarks on weakly-supervised classification and localization of common thorax diseases. In: {CVPR}. pp. 3462--3471. {IEEE} Computer Society (2017)

\bibitem{medclip}
Wang, Z., Wu, Z., Agarwal, D., Sun, J.: Medclip: Contrastive learning from unpaired medical images and text. In: {EMNLP}. pp. 3876--3887. Association for Computational Linguistics (2022)

\bibitem{TCP2024}
Yao, H., Zhang, R., Xu, C.: {TCP:} textual-based class-aware prompt tuning for visual-language model. CoRR  \textbf{abs/2311.18231} (2023)

\bibitem{kgcoop}
Yao, H., Zhang, R., Xu, C.: Visual-language prompt tuning with knowledge-guided context optimization. In: {CVPR}. pp. 6757--6767. {IEEE} (2023)

\bibitem{ZhangXPLVLSHLSX21}
Zhang, J., Xie, Y., Pang, G., Liao, Z., Verjans, J., Li, W., Sun, Z., He, J., Li, Y., Shen, C., Xia, Y.: Viral pneumonia screening on chest x-rays using confidence-aware anomaly detection. {IEEE} Trans. Medical Imaging  \textbf{40}(3),  879--890 (2021)

\bibitem{kad}
Zhang, X., Wu, C., Zhang, Y., Xie, W., Wang, Y.: Knowledge-enhanced visual-language pre-training on chest radiology images  \textbf{14} (2023)

\bibitem{ZhangZYLC21}
Zhang, X., Zhang, Q., Yan, Z., Liu, R., Cao, Y.: Enhancing label correlation feedback in multi-label text classification via multi-task learning. In: {ACL/IJCNLP} (Findings). Findings of {ACL}, vol. {ACL/IJCNLP} 2021, pp. 1190--1200. Association for Computational Linguistics (2021)

\bibitem{convirt}
Zhang, Y., Jiang, H., Miura, Y., Manning, C.D., Langlotz, C.P.: Contrastive learning of medical visual representations from paired images and text. In: {MLHC}. Proceedings of Machine Learning Research, vol.~182, pp. 2--25. {PMLR} (2022)

\bibitem{maskclip}
Zhou, C., Loy, C.C., Dai, B.: Extract free dense labels from {CLIP}. In: {ECCV} {(28)}. Lecture Notes in Computer Science, vol. 13688, pp. 696--712. Springer (2022)

\bibitem{cocoop}
Zhou, K., Yang, J., Loy, C.C., Liu, Z.: Conditional prompt learning for vision-language models. In: {CVPR}. pp. 16795--16804. {IEEE} (2022)

\bibitem{coop}
Zhou, K., Yang, J., Loy, C.C., Liu, Z.: Learning to prompt for vision-language models. Int. J. Comput. Vis.  \textbf{130}(9),  2337--2348 (2022)

\end{thebibliography}

%
%

\newpage
\appendix
\section{Dataset Details}
\subsection{Public Dataset}
\subsubsection{MIMIC-CXR}contains 377,110 chest radiographs and 227,827 associated free-text radiology reports. We used the entire dataset to fine-tune our CLIP backbone.
\subsubsection{CheXpert}contains 224,316 chest radiographs labelled into 14 pathology categories. All labels in the training set were extracted by labeller from the original reports and contain annotations of \(\{1,0,-1\}\), indicating positive, negative, and uncertain. The official validation set and test set are labelled by experienced radiologists. We used the training set for prompt learning phase, the validation set for the best epoch selection and the test set for zero-shot evaluation. In prompt learning, we skipped uncertain CheXpert labels.
\subsubsection{VinDr-CXR} contains 18,000 chest radiographs that were annotated by experienced radiologists with 22 local labels and 6 global labels. The released dataset is divided into a training set of 15,000 and a test set of 3,000. In the multi-label zero-shot evaluation, we used its test set and evaluated all 28 pathology categories.
\subsubsection{PadChest}contains 160,868 chest radiographs labelled with 174 different radiographic findings, 19 differential diagnoses. 39,053 samples were anootated by radiologists. We only used radiologist-labeled subset for zero-shot evaluation and filter classes in which the number of positive samples \(n_{pos} > 50\). 
\subsubsection{Chestxray-14} contains 112,120 images across 15 categories (1 no finding, 14 pathologies) labeled by NLP tools.It was excluded from zero-shot evaluation due to non-radiologist labeling but included in cross-dataset generalization experiments.
\subsection{Processing Details and Statistics}
\subsubsection{Pre-processing.} For MIMIC-CXR, we trimmed report texts to which only include the ``IMPRESSION" or ``CONCLUSION" fields and omitted any physician identifiers or view positions sequences. For all datasets, images were resized to 224 × 224 and normalized using the train dataset's mean and standard deviation. We randomly apply two of these augmentations: \textit{RandomResizedCrop} for crop scale from 0.8 to 1.0; \textit{RandomRotation} for rotate degrees from -10 to 10; \textit{RandomAffine} for affine degrees from -10 to 10, affine translate of 0.0625, affine scale from 0.9 to 1.1, \textit{RandomBrightnessContrast} with brightness from 0.8 to 1.2, contrast from 0.8 to 1.2. And individually set the probability of \textit{RandomHorizontalFlip} to 0.5.

\subsubsection{Metadata.} In Table \ref{tablef1}, We present the metadata of the dataset that has been filtered by our filtering strategy.

\begin{table}[ht]
\centering
\caption{\textbf{Metadata of Datasets.} (*) denotes datasets where the original DICOM version lacks annotations. \((^\dag)\) denotes datasets containing \textbf{image, label, and text} sources, but \textbf{only image-label} pairs were utilized for our evaluation.  \((^\star)\) denotes the exclusion of a category (``Edema" in VinDr-CXR-test) due to the absence of positive samples. }
\label{tablef1}
\begin{tabularx}{\textwidth}{@{}lXXXX@{}}
\toprule
\textbf{Dataset} & Filtered Data & Categories & Type & Annotations \\ 
\midrule
MIMIC-CXR       & 203,394                    & --                  & Image-Text                & No Annotation*                  \\
\(\text{CheXpert}_\text{train}\)       & 191,027                     & 14                 & Image-Label                  & NLP               \\
\(\text{CheXpert}_\text{val}\)      & 202                     & 14                  & Image-Label                  & Radiologists               \\
\(\text{CheXpert}_\text{test}\)       & 518                     & 14                  & Image-Label                  &   Radiologists             \\
\(\text{VinDr-CXR}_\text{test}\)        & 3,000                    & 28 (\(27^\star\))                  & Image-Label                 & Radiologists          \\ 
\(\text{ChestXray}_\text{train}\)      & 86,524                   & 15                  & Image-Label                 & NLP          \\ 
\(\text{ChestXray}_\text{test}\)         & 25,596                    & 15                  & Image-Label                 & NLP          \\ 
\(\text{PadChest}_\text{test}\)        & 26,432                    & 108 (\(n_{pos}>50\))                 & Image-Label-Text\(^\dag\)                 & Radiologists         \\
Private-CXR        & 7,357                    & 18                  & Image-Label-Text\(^\dag\)          & Radiologists         \\ 

\bottomrule
\end{tabularx}
\end{table}
\begin{figure}[htbp]
\centering
\includegraphics[width=\textwidth]{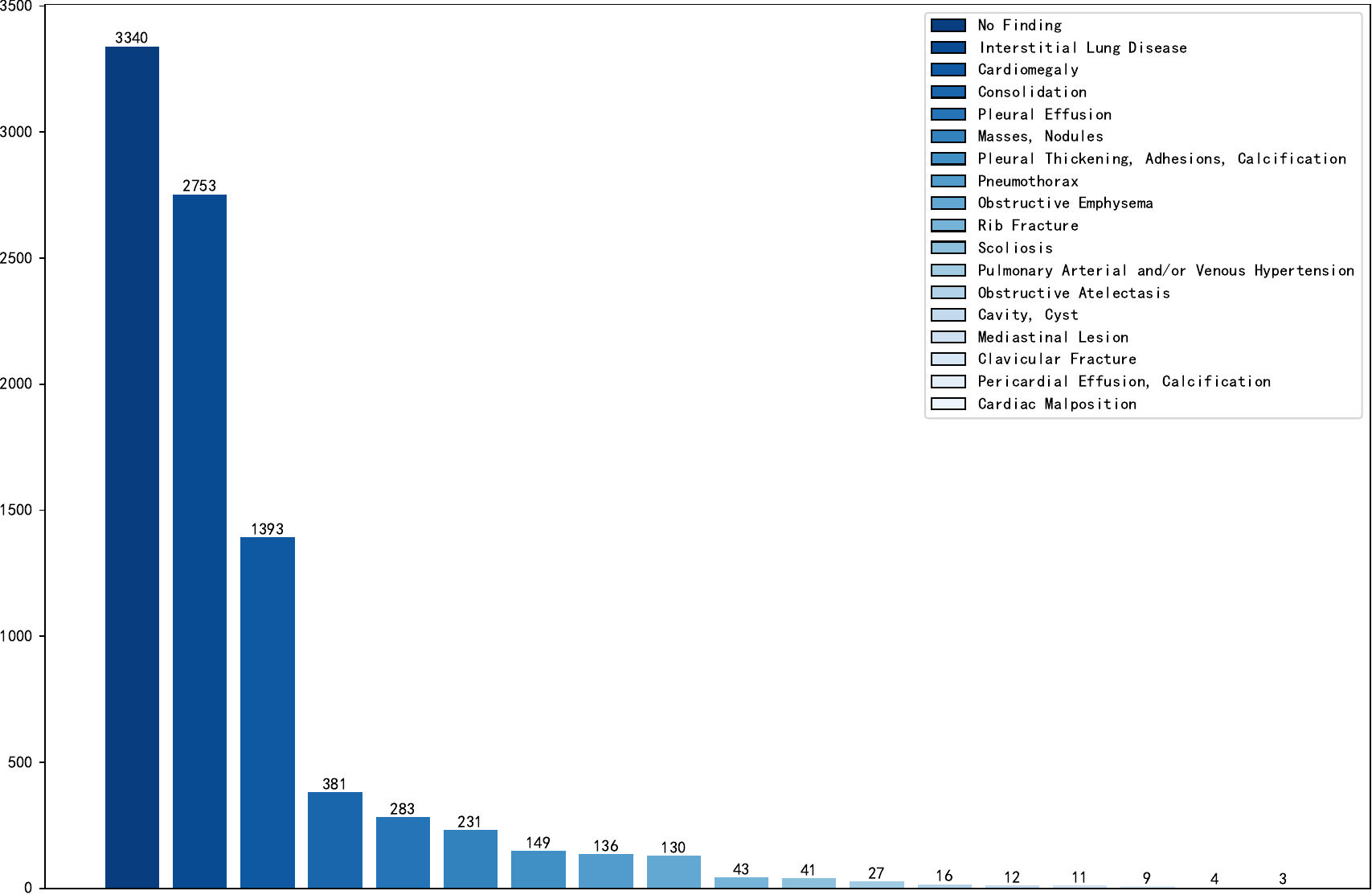}
\caption{\textbf{Detailed Statistics about our Private-CXR.} The subfigure in the upper left corner lists the names of each category. }

\label{fig4}
\end{figure}
\subsection{Private Dataset}
We present the detailed statistics of our Private-CXR in Fig. \ref{fig4}. Private-CXR includes 18 pathology categories, in which 13 are unseen in CheXpert's annotation strategy. In all 7357 samples, 16.2\% (1191) are multi-label and 83.8\% (6166) are single-label. 
\clearpage
\section{Supplementary Experiments}
\subsection{Generalization Study on Cross-Dataset Training}
While merging multi-label datasets with varied categories for scaling training is complex, fine-tuning an open-vocabulary method across datasets is straightforward. We conducted cross-dataset experiments on ChestXray-14 and CheXpert. \( \text{CheXpert} \rightarrow \text{ChestXray}\) in Table \ref{tablef2} denotes training on CheXpert followed by fine-tuning on ChestXray. We compare the generalizability with the existing state-of-the-art multi-label prompt learning method, DualCoOp.
\begin{table}[ht]
\centering
\caption{\textbf{Cross-Dataset Training on CheXpert and ChestXray-14.} Besides, we use VinDr-CXR and Private-CXR for generalization evaluation. \textbf{PsPG} is highlighted with \colorbox{pink!20}{background}. * means performing evaluation in the domain of the last train dataset.}
\label{tablef2}
\resizebox{\textwidth}{!}{
\begin{tabular}{@{}lcccccccccccccccc@{}}
\toprule
    \multirow{3}{*}{\textbf{Train Dataset}}  & \multicolumn{4}{c}{\textbf{CheXpert}} & \multicolumn{4}{c}{\textbf{ChestXray-14}} & \multicolumn{4}{c}{\textbf{Private-CXR}} & \multicolumn{4}{c}{\textbf{VinDr-CXR}}\\
\cmidrule(lr){2-5} \cmidrule(lr){6-9}\cmidrule(lr){10-13} \cmidrule(lr){14-17}
  & \mysmallscale{\makecell{Macro\\AUC}} & \mysmallscale{\makecell{Micro\\AUC}} & \mysmallscale{mAP} &  &  \mysmallscale{\makecell{Macro\\AUC}} & \mysmallscale{\makecell{Micro\\AUC}} & \mysmallscale{mAP} &  &  \mysmallscale{\makecell{Macro\\AUC}} & \mysmallscale{\makecell{Micro\\AUC}} & \mysmallscale{mAP}&  & \mysmallscale{\makecell{Macro\\AUC}} & \mysmallscale{\makecell{Micro\\AUC}} & \mysmallscale{mAP} &  \\
\midrule
  & 85.0* & 85.4* & 47.5* &  & \textbf{67.7} & \textbf{67.5} & \textbf{18.0} &  & 69.9 & 58.0 & 18.5 & & 72.2 & 69.8 & 18.6 &\\
\rowcolor{pink!20}
 \cellcolor{white}\multirow{-2}{*}{Only CheXpert} & 85.2* & 85.1* & 49.3* &  & 61.8 & 63.9 & 16.8 &  & \textbf{77.9} & \textbf{80.5} & \textbf{18.8} & & \textbf{75.1} & \textbf{78.4} & \textbf{19.9} &\\
\midrule
  & 70.9 & 51.0 & 35.2 &  & 75.1* & 84.4* & 22.0* &  & \textbf{72.8} & 65.8 & 14.7 & & 70.9 & 80.2 & \textbf{20.4} &\\
\rowcolor{pink!20}
 \cellcolor{white}\multirow{-2}{*}{Only ChestXray} & \textbf{72.0} & \textbf{67.4} & \textbf{37.5} &  & 75.0* & 84.3* & 22.0* &  & 72.1 & \textbf{77.9} & \textbf{16.1} & & \textbf{74.1} & \textbf{80.7} & 16.7&\\
\midrule
  & 78.4 & \textbf{76.3} & 41.8 &  & 75.7* & 84.5* & 22.7* &  & 68.3 & 67.4 & \textbf{17.9} & & 75.5 & 78.9 & 19.3 &\\
\rowcolor{pink!20}
 \cellcolor{white}\multirow{-2}{*}{\( \text{CheXpert} \rightarrow \text{ChestXray}\)} & \textbf{82.8} & 75.6 & \textbf{45.2} &  & 75.7* & 84.5* & 22.3* &  & \textbf{77.1} & \textbf{81.0} & 17.1 & & \textbf{82.4} & \textbf{81.5} & \textbf{22.3} &\\
\midrule
  & 85.2* & 86.0* & 48.6* &  & 71.8 & 72.9 & \textbf{19.8} &  & \textbf{76.8} & 65.8 & 18.8 & & \textbf{76.4} & 73.4 & \textbf{22.5} &\\
\rowcolor{pink!20}
 \cellcolor{white}\multirow{-2}{*}{\( \text{ChestXray} \rightarrow \text{CheXpert}\)} & 85.2* & 85.5* & 47.0* &  & \textbf{72.2} & \textbf{75.0} & \textbf{19.8} &  & 73.1 & \textbf{77.8} & \textbf{19.3} & & 75.3 & \textbf{76.9} & 19.5 &\\
\bottomrule
\end{tabular}
}
\end{table}
\subsubsection{Result.} In zero-shot scenarios for out-of-domain categories, our PsPG method surpasses DualCoOp across numerous metrics, illustrating its cross-dataset generalizability and the potential for application to broader datasets. 

Given that larger medical multi-label datasets often rely on NLP for annotations, our training and fine-tuning attempts are all performed on datasets with potential label noise. PsPG achieves fairly high zero-shot performance relative to MedCLIP, which uses CheXpert's labeller to construct sentence-level annotations. We speculate that the indirect nature of prompt learning may mitigate the impact of label noise to a certain degree.

\subsection{Sequence Length Ablation}
We conducted sequence length ablation study. The result is shown in Table \ref{tablef3}. We observed the impact of sequence length on model performance; specifically, a sequence length of 8 slightly underperforms compared to 16. Conversely, a length of 32, which extends the autoregressive process duration, yields poorer results than both shorter lengths. Moreover, longer sequences seems to reduce Macro AUC on CheXpert, though Micro AUC surpassed the 16-length setting. These findings imply that shorter prompt sequences might be more optimal for our prompt learning approach.
\begin{table}[ht]
\centering
\caption{This table includes Sequence Length Ablation. We show the average above, and 95\% CI as (\( \cdot  , \cdot  \)).}
\label{tablef3}
\begin{tabular}{@{}lcccccccccc@{}}
\toprule
    {\textbf{Seq.}}  & \multicolumn{3}{c}{\textbf{CheXpert}} & \multicolumn{3}{c}{\textbf{VinDr-CXR}} & \multicolumn{3}{c}{\textbf{Private-CXR}} &\\
    \cmidrule(lr){2-4} \cmidrule(lr){5-7}\cmidrule(lr){8-10}
    & \mysmallscale{Macro AUC} & \mysmallscale{Micro AUC} &   &  \mysmallscale{Macro AUC} & \mysmallscale{Micro AUC} &   &  \mysmallscale{Macro AUC} & \mysmallscale{Micro AUC} &  & \\
\midrule
Length 8 & \makecell{84.5\\(82.6, 86.4)}  & \makecell{83.9\\(83.1, 84.6)} &  & \makecell{73.5\\(71.2, 75.5)} & \makecell{74.4\\(73.4, 75.8)} &  & \makecell{74.8\\(72.2, 77.2)} & \makecell{76.1\\(75.7, 76.8)}&\\
\textbf{Length 16} & \makecell{85.2\\(82.6, 87.7)} & \makecell{85.1\\(84.2, 85.9)} & & \makecell{\textbf{75.1}\\(72.0, 77.7)} & \makecell{\textbf{78.4}\\(77.4, 79.4)} &   & \makecell{\textbf{77.9}\\(75.4, 80.0)} & \makecell{\textbf{80.5}\\(80.0, 80.9)} &\\
Length 32 & \makecell{\underline{80.6}\\(76.8, 83.8)} & \makecell{85.4\\(84.3, 86.1)} &  & \makecell{69.9\\(67.5, 72.4)} & \makecell{72.6\\(71.6, 73.5)} &   & \makecell{68.4\\(65.9, 71.4)} & \makecell{68.4\\(67.9, 68.9)} &\\
\bottomrule
\end{tabular}
\end{table}

\subsection{Influence of Batch Size in Evaluation}
Our proposed structure, as detailed in \ref{pd}, utilizes global image features within a mini-batch as the visual input for the \textit{Prompt Decoder}. This implies that the batch size during evaluation may influence performance outcomes. So we varied the batch size and conducted evaluations, with findings summarized in Table \ref{tablef4}. These results indicate that the influence of different batch sizes during evaluation on our method is minimal. (typically \textless1\% for average results).
\begin{table}[ht]
\centering
\caption{This table includes experiments with different batch size during evaluation. We show the average above, and 95\% CI as (\( \cdot  , \cdot  \)).}
\label{tablef4}
\begin{tabular}{@{}ccccccccccc@{}}
\toprule
     \multirow{2}{*}{\makecell{\textbf{Val.} \\ \textbf{Batch Size}}}  & \multicolumn{3}{c}{\textbf{CheXpert}} & \multicolumn{3}{c}{\textbf{VinDr-CXR}} & \multicolumn{3}{c}{\textbf{Private-CXR}} &\\
    \cmidrule(lr){2-4} \cmidrule(lr){5-7}\cmidrule(lr){8-10}
    & \mysmallscale{Macro AUC} & \mysmallscale{Micro AUC} &   &  \mysmallscale{Macro AUC} & \mysmallscale{Micro AUC} &   &  \mysmallscale{Macro AUC} & \mysmallscale{Micro AUC} &  & \\
\midrule
16 & \makecell{85.2\\(82.5, 87.7)}  & \makecell{85.1\\(84.2, 85.9)} &  & \makecell{75.1\\(72.1, 77.8)} & \makecell{78.4\\(77.4, 79.4)} &  & \makecell{77.9\\(75.4, 80.0)} & \makecell{80.5\\(80.0, 80.9)}&\\
32 & \makecell{85.3\\(82.7, 87.8)}  & \makecell{85.1\\(84.3, 85.9)} &  & \makecell{75.1\\(72.1, 77.8)} & \makecell{78.4\\(77.4, 79.3)} &  & \makecell{77.9\\(75.4, 79.9)} & \makecell{80.5\\(80.0, 80.9)}&\\
\textbf{64} & \makecell{85.2\\(82.6, 87.7)} & \makecell{85.1\\(84.2, 85.9)} & & \makecell{75.1\\(72.0, 77.7)} & \makecell{78.4\\(77.4, 79.4)} &   & \makecell{77.9\\(75.4, 80.0)} & \makecell{80.5\\(80.0, 80.9)} &\\
128 & \makecell{85.2\\(82.5, 87.7)} & \makecell{85.1\\(84.2, 85.8)} &  & \makecell{75.1\\(72.0, 77.7)} & \makecell{78.4\\(77.4 79.3)} &   & \makecell{77.9\\(75.4, 79.9)} & \makecell{80.5\\(80.0, 81.1)} &\\
\bottomrule
\end{tabular}
\vspace{-10px}
\end{table}
\subsection{Detailed Confidence Intervals}
\begin{table}[htbp]
\centering
\caption{We show the average above, and 95\% CI as (\( \cdot  , \cdot  \)) for Multi-label Zero-shot Classification in \ref{mlzc}.}
\label{tablef5}
\begin{tabular}{@{}lccccccccc@{}}
\toprule
    {\textbf{Method}}  & \multicolumn{4}{c}{\textbf{CheXpert}} & \multicolumn{4}{c}{\textbf{VinDr-CXR}}&\\
    \cmidrule(lr){2-5} \cmidrule(lr){6-9} 
    & \mysmallscale{Macro AUC} & \mysmallscale{Micro AUC} & \mysmallscale{mAP} & &  \mysmallscale{Macro AUC} & \mysmallscale{Micro AUC} & \mysmallscale{mAP} &  &  \\
\midrule
ConVIRT & \makecell{70.7\\(67.1, 74.9)}  & \makecell{66.2\\(64.8, 67.9)} & \makecell{30.3\\(27.9, 33.2)}& & \makecell{72.7\\(69.9, 74.5)}  & \makecell{71.7\\(71.1, 72.5)} & \makecell{13.0\\(12.2, 13.7)}&\\
GLoRIA & \makecell{75.9\\(72.6, 79.1)}  & \makecell{73.2\\(72.2, 74.2)} & \makecell{42.3\\(39.5, 46.3)}& & \makecell{64.2\\(62.3, 66.2)}  & \makecell{51.9\\(50.9, 53.6)} & \makecell{13.8\\(11.2, 16.4)}&\\
MedCLIP &\makecell{76.6\\(72.2, 80.1)}  & \makecell{83.4\\(82.3, 84.6)} & \makecell{47.7\\(42.5, 53.1)}& & \makecell{58.9\\(56.3, 61.1)}  & \makecell{71.6\\(71.2, 72.3)} & \makecell{22.4\\(19.8, 24.1)}&\\
BioViL-T & \makecell{70.6\\(68.4, 73.9)}  & \makecell{67.6\\(66.0, 69.7)} & \makecell{36.8\\(34.1, 41.0)}& & \makecell{81.2\\(79.3, 82.8)}  & \makecell{72.2\\(71.3, 72.8)} & \makecell{19.3\\(18.6, 20.9)}&\\
Chexzero & \makecell{72.7\\(68.1, 75.9)}  & \makecell{60.6\\(59.6, 61.9)} & \makecell{41.6\\(38.0, 45.5)}& & \makecell{75.9\\(73.5, 78.5)}  & \makecell{62.6\\(62.0, 63.4)} & \makecell{19.6\\(17.8, 22.8)}&\\
\(\text{CLIP}_{\text{fine-tuned}}\) & \makecell{74.4\\(70.5, 78.2)}  & \makecell{69.8\\(68.2, 71.1)} & \makecell{42.0\\(39.1, 44.6)}& & \makecell{75.0\\(72.9, 76.8)}  & \makecell{71.8\\(70.9, 72.5)} & \makecell{18.5\\(17.1, 20.1)}&\\
KAD & \makecell{83.8\\(79.0, 86.8)}  & \makecell{85.2\\(84.3, 86.3)} & \makecell{50.6\\(46.2, 54.1)}& & \makecell{82.6\\(80.5, 84.0)}  & \makecell{82.4\\(81.9, 83.2)} & \makecell{24.7\\(22.7, 26.7)}&\\
\midrule
\(\text{DualCoOp}_{\text{init}}\) & \makecell{83.6\\(80.7, 86.5)}  & \makecell{84.6\\(83.7, 85.3)} & \makecell{46.3\\(40.7, 50.6)}& & \makecell{72.9\\(70.5, 75.9)}  & \makecell{74.7\\(74.2, 75.4)} & \makecell{19.6\\(17.9, 21.3)}&\\
\(\text{DualCoOp}_{\text{16}}\)& \makecell{85.0\\(82.1, 87.6)}  & \makecell{85.4\\(84.6, 86.0)} & \makecell{47.5\\(41.8, 51.6)}& & \makecell{72.2\\(69.4, 76.1)}  & \makecell{69.8\\(68.6, 70.6)} & \makecell{18.6\\(16.8, 20.2)}&\\
\(\text{PsPG}_{\text{prefix}}\) & \makecell{84.8\\(82.4, 87.2)}  & \makecell{84.8\\(83.8, 85.5)} & \makecell{46.4\\(42.0, 49.1)}& & \makecell{67.8\\(64.9, 71.8)}  & \makecell{75.2\\(74.2, 76.0)} & \makecell{18.2\\(16.0, 19.7)}&\\
\(\text{PsPG}\) & \makecell{85.2\\(82.6, 87.7)}  & \makecell{85.1\\(84.2, 85.9)} & \makecell{49.3\\(42.4, 54.3)}& & \makecell{75.1\\(72.0, 77.7)}  & \makecell{78.4\\(77.4, 79.4)} & \makecell{19.9\\(18.1, 21.4)}&\\

\\
\toprule
    {\textbf{Method}}  & \multicolumn{4}{c}{\textbf{Private-CXR}} & \multicolumn{4}{c}{\textbf{PadChest}}&\\
    \cmidrule(lr){2-5} \cmidrule(lr){6-9} 
    & \mysmallscale{Macro AUC} & \mysmallscale{Micro AUC} & \mysmallscale{mAP} & &  \mysmallscale{Macro AUC} & \mysmallscale{Micro AUC} & \mysmallscale{mAP} &  &  \\
\midrule
ConVIRT & \makecell{70.6\\(68.7, 72.7)}  & \makecell{64.3\\(63.8, 64.8)} & \makecell{11.6\\(11.1, 12.3)}& & \makecell{63.7\\(63.4, 64.0)}  & \makecell{58.5\\(58.4, 58.6)} & \makecell{3.6\\(3.6, 3.6)}&\\
GLoRIA & \makecell{76.9\\(75.2, 78.4)}  & \makecell{56.7\\(56.3, 57.2)} & \makecell{16.2\\(15.0, 18.2)}& & \makecell{67.3\\(67.3, 67.3)}  & \makecell{59.3\\(59.2, 59.3)} & \makecell{4.5\\(4.4, 4.5)}&\\
MedCLIP & \makecell{57.2\\(54.8, 58.8)}  & \makecell{71.1\\(70.6, 71.7)} & \makecell{18.0\\(17.3, 19.0)}& & \makecell{54.7\\(54.6, 54.9)}  & \makecell{54.5\\(54.5, 54.6)} & \makecell{5.3\\(5.3, 5.4)}&\\
BioViL-T & \makecell{76.0\\(74.7, 76.7)}  & \makecell{61.6\\(61.3, 61.8)} & \makecell{14.8\\(14.5, 15.5)}& & \makecell{69.2\\(69.0, 69.6)}  & \makecell{66.9\\(66.8, 67.0)} & \makecell{5.3\\(5.2, 5.4)}&\\
Chexzero & \makecell{72.1\\(69.8, 74.1)}  & \makecell{71.1\\(70.7, 71.5)} & \makecell{18.6\\(17.0, 20.9)}& & \makecell{65.8\\(65.8, 65.9)}  & \makecell{65.7\\(65.5, 65.8)} & \makecell{5.9\\(5.9, 6.0)}&\\
\(\text{CLIP}_{\text{fine-tuned}}\) & \makecell{76.3\\(74.0, 78.3)}  & \makecell{72.4\\(72.0, 72.9)} & \makecell{16.5\\(15.8, 17.6)}& & \makecell{69.9\\(69.6, 70.1)}  & \makecell{67.2\\(67.2, 67.3)} & \makecell{6.4\\(6.3, 6.5)}&\\
KAD & \makecell{77.8\\(75.5, 80.1)}  & \makecell{70.1\\(69.5, 70.9)} & \makecell{20.7\\(19.7, 21.9)}& & \makecell{75.7\\(75.4, 75.8)}  & \makecell{79.1\\(79.0, 79.2)} & \makecell{10.1\\(9.9, 10.2)}&\\
\midrule
\(\text{DualCoOp}_{\text{init}}\) & \makecell{67.5\\(64.3, 70.82)}  & \makecell{51.1\\(50.7, 51.5)} & \makecell{19.6\\(18.6, 20.7)}& & \makecell{65.7\\(65.5, 65.8)}  & \makecell{67.1\\(66.9, 67.2)} & \makecell{7.3\\(7.1, 7.5)}&\\
\(\text{DualCoOp}_{\text{16}}\)& \makecell{69.9\\(66.6, 72.2)}  & \makecell{58.0\\(57.5, 58.7)} & \makecell{18.5\\(17.7, 19.9)}& & \makecell{64.0\\(63.7, 64.8)}  & \makecell{63.5\\(63.5, 63.6)} & \makecell{6.1\\(6.0, 6.3)}&\\
\(\text{PsPG}_{\text{prefix}}\) & \makecell{67.7\\(65.4, 69.7)}  & \makecell{66.9\\(66.4, 67.5)} & \makecell{18.4\\(17.7, 19.5)}& & \makecell{60.8\\(60.7, 61.0)}  & \makecell{63.0\\(62.9, 63.1)} & \makecell{4.3\\(4.3, 4.4)}&\\
\(\text{PsPG}\) & \makecell{77.9\\(75.4, 80.0)}  & \makecell{80.5\\(80.0, 80.9)} & \makecell{18.8\\(18.0, 20.2)}& & \makecell{62.2\\(62.2, 62.3)}  & \makecell{61.6\\(61.5, 61.6)} & \makecell{4.0\\(4.0, 4.0)}&\\

\bottomrule
\end{tabular}

\end{table}

\end{document}